\documentclass[10pt,twocolumn,letterpaper]{article}

\usepackage[pagenumbers]{cvpr} 

\usepackage{graphicx}
\usepackage{amsmath}
\usepackage{amssymb}
\usepackage{booktabs}
\usepackage[symbol]{footmisc}

\newcommand{\rebut}[1]{#1}
\newcommand{\figref}[1]{Fig.~\ref{#1}}
\newcommand{\secref}[1]{Section \ref{#1}}

\usepackage[pagebackref,breaklinks,colorlinks]{hyperref}

\usepackage[capitalize]{cleveref}
\crefname{section}{Sec.}{Secs.}
\Crefname{section}{Section}{Sections}
\Crefname{table}{Table}{Tables}
\crefname{table}{Tab.}{Tabs.}

\def\cvprPaperID{95} % *** Enter the CVPR Paper ID here
\def\confName{ICDL}
\def\confYear{2023}

\begin{document}

\title{Caregiver Talk Shapes Toddler Vision: A Computational Study of Dyadic Play}

\author{Timothy Schaumlöffel$^{1,2,*}$, Arthur Aubret$^{3,4,*}$, Gemma Roig$^{1,2,\dag}$, Jochen Triesch$^{1,3,\dag}$ \\
$^{1}$Goethe University, Frankfurt am Main, Germany\\
$^{2}$The Hessian Center for Artificial Intelligence (hessian.AI), Darmstadt, Germany\\
$^{3}$Frankfurt Institute for Advanced Studies, Frankfurt am Main, Germany\\
$^{4}$Xidian-FIAS international Joint Research Center, Frankfurt am Main, Germany\\
{\tt\small \{schaumloeffel, roig\}@em.uni-frankfurt.de} \\
{\tt\small \{aubret, triesch\}@fias.uni-frankfurt.de} 
}
\maketitle
\footnotetext[1]{Equal contribution}
\footnotetext[2]{Shared last authorship}
%%%%%%%%% ABSTRACT
\begin{abstract}
Infants' ability to recognize and categorize objects develops gradually. The second year of life is marked by both the emergence of more semantic visual representations and a better understanding of word meaning. This suggests that language input may play an important role in shaping visual representations. However, even in suitable contexts for word learning like dyadic play sessions, caregivers utterances are sparse and ambiguous, often referring to objects that are different from the one to which the child attends. Here, we systematically investigate to what extent caregivers' utterances can nevertheless enhance visual representations. \rebut{For this we propose a computational model of visual representation learning during dyadic play}. We introduce a synthetic dataset of ego-centric images perceived by a toddler-agent that moves and rotates toy objects in different parts of its home environment while ``hearing" caregivers' utterances, modeled as captions. We propose to model toddlers' learning as simultaneously aligning representations for 1) close-in-time images and 2) co-occurring images and utterances. We show that utterances with statistics matching those of real caregivers give rise to representations supporting improved category recognition. Our analysis reveals that a small decrease/increase in object-relevant naming frequencies can drastically impact the learned representations. This affects the attention on object names within an utterance, which is required for efficient visuo-linguistic alignment. Overall, our results support the hypothesis that caregivers' naming utterances can improve toddlers' visual representations. Code and dataset are available at: \url{https://github.com/neuroai-arena/ToddlerVisionLearning.git}
\end{abstract}

\begin{figure*}
    \centering
    \includegraphics[width=1\linewidth]{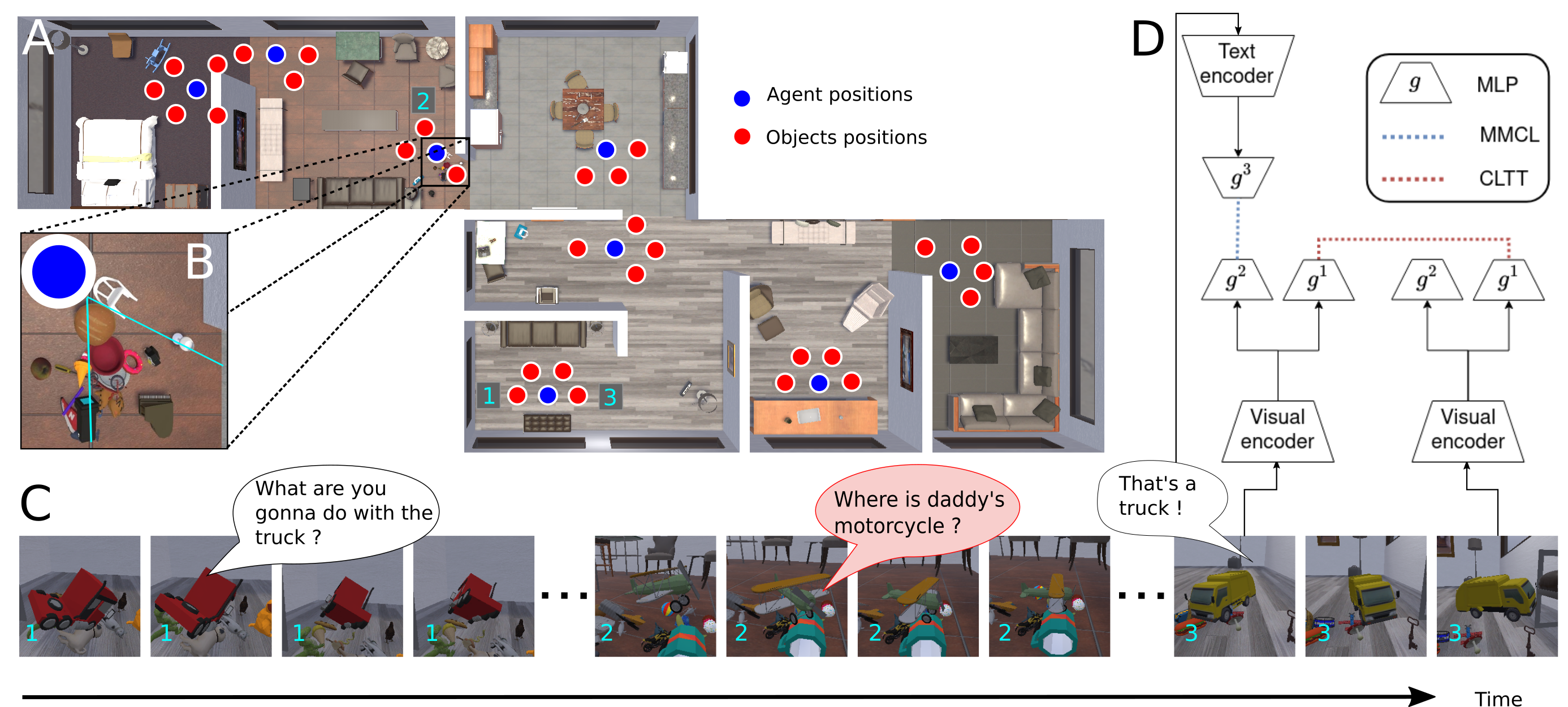}
    \caption{A) Top view of the \textit{Virtual Home Environment}, where blue and red dots, respectively, indicate possible agent and toy positions. The agent is always turned towards a toy position. Toy positions marked 1--3 correspond to sessions 1--3 in C. B) Zoom-in of the scene in A with turquoise lines indicating the agent's \rebut{field of view}. C)  Images extracted in a temporally ordered fashion (left to right) for three different ``play" sessions of the Dyadic Play Dataset. Text boxes show examples of captions related to the manipulated object (white) or to another object in the background (red). D) Summary of the learning architecture, see \secref{sec:ssl} for details. Abbreviations: MLP: multi-layer perceptron, MMCL: multimodal contrastive learning, CLTT: contrastive learning through time.}
    \label{fig:dyadicfull}
\end{figure*}

\section{Introduction}
\label{sec:intro}
% Todllers are good learners presumably because of time and language
 Second-year toddlers are proficient learners of object representations that support the recognition of objects independently of viewpoint (object instance recognition) and the assignment of novel exemplars to learned categories (object categorization recognition). What learning mechanisms support this development? On one side, there is evidence that biological organisms learn similar representations for close-in-time visual inputs \cite{wood2018development,li2010unsupervised}. This is called the \textit{slowness principle} \cite{wiskott2002slow} and it supports the building of view-invariant object representations during, e.g., object manipulations \cite{kraebel2006three,soska2008development}. On the other side, the acquisition of object names correlates with a stronger focus on global shape features relative to texture \cite{gershkoff2004shape,smith2002object,jones2003late}, which spurs performance in name-agnostic categorization tasks \cite{diesendruck2003specific,nelson2000young}, see also \cite{gopnik1987development}. For learning object names, the synchronization of toddlers' visual attention on the object and its naming by the caregivers is crucial \cite{schroer2023looking}.

Dyadic play could be particularly important for promoting the emergence of more semantic object representations, because both learning mechanisms are likely to co-occur during such play: the toddler extensively focuses on/manipulates objects and the caregiver frequently talks to the toddler about the present objects \cite{suarez2022joint,schroer2022visual}.
However, naming utterances during such play are \textit{sparse} \cite{clerkin2022real} and \textit{ambiguous}: they typically contain several words unrelated to the object. Furthermore, even though toddlers have biased attention towards held objects \cite{yang2023using}, they often play in contexts that admit several other objects in the background \cite{bambach2016active,yu2021infant}, adding another potential source of confusion.

In this paper, we investigate two what extent dyadic play sessions can support the emergence of semantic object representations in toddlers. \rebut{We tackle this question by proposing a computational model of toddlers' learning during dyadic play}. First, we introduce the Dyadic Play Dataset, a novel dataset of images simulating at-home ego-centric dyadic play sessions. In this dataset, \rebut{we simulate a playing toddler} that moves and turns 3D toy models extracted from the Toy4k dataset \cite{stojanov2021using}. \rebut{The agent ``plays"} in front of a background composed of household furniture (cf. \figref{fig:dyadicfull}A) and a plausible number of toys scattered on the floor (cf. \figref{fig:dyadicfull}B). As shown in \figref{fig:dyadicfull}C, The agent occasionally ``hears" caregiver's utterances, modeled as captions \rebut{extracted from the CHILDES database, a comprehensive data repository of children's language acquisition including child-directed speech \cite{macwhinney1985child}}. Then, we consider a bio-inspired model of toddlers' learning that 1) maps close-in-time visual inputs to similar representations and 2) similarly aligns the representations of visual inputs and co-occurring naming utterances (\figref{fig:dyadicfull}D). \rebut{We further simulate toddlers' visual attention by biasing \rebut{the model} to extract \rebut{visual} features from the currently held object.} The overall model allows us to systematically study the potential impact of the sparsity and ambiguity of naming utterances on the learned visual representations. 

Our experiments show that utterance statistics reported in developmental psychology experiments support the construction of semantic visual representations. Our analysis shows that realistic object-relevant naming frequencies fall within a range of values \rebut{for which a multiplicative factor of two} drastically impact the learned representation. This impacts whether the model attends to the object name within an utterance, which is necessary to \rebut{efficiently} guide visual representations.  
Thus, our paper provides computational support to the hypothesis that caregivers' \rebut{sparse and ambiguous} utterances help to build visual representations during dyadic play sessions.

\section{Related work}

\paragraph{Models of toddlers' object learning}

Recently, the availability of datasets of images extracted from toddlers' head-mounted cameras \cite{bambach2016active,sullivan2021saycam} \rebut{has enabled} the study of representations learnt through \rebut{the} senses of toddlers. This way, two recent models managed to learn word-vision mappings, but they either used a pretrained vision model \cite{vong2021cross,tsutsui2020computational} or trained on curated data in a supervised fashion \cite{tsutsui2020computational}. Another model \rebut{extracted} semantic visual representations by making similar close-in-time representations \cite{orhan2020self}, but it \rebut{did} not \rebut{leverage} any sort of utterances. Working with such real-world ego-centric data has the limitation that a precise control of the statistics of the training data, like the sparsity or ambiguity of naming utterances, is difficult. This makes it \rebut{hard} to study how such statistics might affect toddlers' object learning. \rebut{A recent model} also trained visual representations in cross-situational contexts, \rebut{\textit{i.e.} when} naming utterances can refer to several \rebut{visible} objects \cite{vong2022cross}. However, they only consider synthetic images based on 9 digits rather than simulated ego-centric play sessions.

\paragraph{Slowness principle for learning visual representations}

Several computational models showed that making similar close-in-time representations can lead to visual representation suitable for object instance recognition, scene recognition and object categorization \cite{franzius2011invariant, schneider2021contrastive, feichtenhofer2021large,aubrettime}. Integrating constraints from toddlers' perceputal experience into these models, like short arms or foveation, can lead to improved ability to recognize objects in front of novel backgrounds \cite{aubret2022toddler}. None of these works modelled interactions with a caregiver whose utterances provide weak language supervision.

\paragraph{Visuo-language representation learning}

Previous works explored multi-modal contrastive learning of audio-visual or text-visual representations and showed that language can supervise visual representation learning \cite{radford2021learning,desai2021virtex,harwath2018jointly}. However, none of these works studied whether naming statistics in dyadic play sessions elicit semantic visual representations. A recent work showed that adding sparse labeling on top of contrastive learning through time enhances visual categorization \cite{aubret2022toddler}. Unlike them, we here consider textured objects in a cluttered home environment and developmentally-relevant statistics of caregivers' utterances.

\section{Methods}

In this section, we expose our computational model of visual representation learning during dyadic play. We first describe the Dyadic Play Dataset, which simulates dyadic play with objects (\secref{sec:dataset}). Then, we explain in \secref{sec:ssl} how we model a toddler's learning process. 

\subsection{Dyadic Play Dataset (DPD)} \label{sec:dataset}

The DPD contains 857,760 images of $224 \times 224$ pixels split into 42,888 ``play sessions," each containing $20$ images. There are 12 sessions dedicated to each of 3574 objects \cite{aubrettime}, \rebut{which are} extracted from the Toys4k dataset \cite{stojanov2021using}. To create them, we simulate \rebut{a toddler} that interacts with objects using the simulation platform ThreeDWorld (TDW) \cite{gan2020threedworld}.

\paragraph{Image recording} At the beginning of each play session, we place the agent in a random location, with a random orientation within the \textit{Virtual Home Environment} \cite{aubrettime} (cf. \figref{fig:dyadicfull}A).
\rebut{We approximate an} ego-centric view of a seated toddler by positioning a camera at $0.4$ m above the ground. \rebut{At the beginning, the camera always watches the object being held, which we call ``main object," making the camera orientation dependent on the initial position of the main object.} \rebut{Because toddlers' short arms constrain the way they hold objects, we randomly sample the relative starting position of the main object within a distance range of $[0.25; 0.35]$ m from the agent, $[-30; 30]^\circ$ on the sides and $[-30; -10]^\circ$ in elevation. Furthermore, toddlers tend to keep objects in an upright position \cite{pereira2010early}, hence, we randomly orient the object} around the yaw axis (unbounded) \rebut{but bound the object orientation in $[-5; 5]^\circ$ around the pitch and roll axes.}

\rebut{In addition to the main object,} we add $5$ to $20$ randomly sampled (without replacement) \rebut{background} objects to the scene, \rebut{following the number of \rebut{toys} commonly present in the field of view of toddlers during play \cite{bambach2016active}.} \rebut{To place them,} we follow the same procedure \rebut{as for the main object}, but change the range of distances from the agent to $[0.42; 0.875]$ m.
\rebut{In contrast to the object being held, we let} them fall to the floor one at a time using the physics engine of TDW. \rebut{\figref{fig:dyadicfull}B shows a top-view of a subsequent play area}.

\rebut{To simulate natural interactions, in} each of the $20$ frames of the play sessions the agent slowly moves and turns the object along/around the three axes. \rebut{We display a high variability of manipulation sequences by generating} all movements with an Ornstein-Uhlenbeck stochastic process \rebut{for which we randomly generate} a new recall coefficient in $[0.1, 1]$ at the beginning of \rebut{each} session, for each kind of manipulation and axis. \rebut{Thus, the agent dynamically changes the manipulation speed within a session, as well as the amplitude of changes from one session to the next, independently for each axis of rotation/motion.} \rebut{We bound} the motion speed by $3^\circ$ per frame for side/elevation motion and $0.05$ m for in-depth motion. However, the agent can not exceed the starting absolute bounds described above. Similarly, we bound the rotation speed by $20^\circ$ around the yaw axis and $4^\circ$ around the pitch and roll axis, also to simulate the upright bias of toddlers \cite{pereira2010early}.

Finally, \rebut{while the agent observes the moving object,} it also executes additional relative small eye movements around the object in focus (yaw and pitch axis), whose endpoints are sampled from a Normal distribution $\mathcal{N}([0,0],2\times\mathbb{I})$. We also allow the agent to focus its gaze on an object different from the main one with probability $0.7$; however, we discard these images during training as previous methods already analyzed the impact of changes of attention on contrastive through time losses \cite{schneider2021contrastive, aubrettime}. \figref{fig:dyadicfull}D shows examples of the resulting sequences of toddler-centric views.

For evaluation purposes, we build a test dataset of $5$ images per main object (a total of 17,870 images), each in different scenes generated as above. To avoid correlated images, we do not apply temporal manipulations used during training. \rebut{Since rotations around pitch/roll axis may go beyond their initial boundaries after some manipulation time during training, it may create a subsequent data distribution shift between train and test images. Assuming that the pitch/roll rotations rarely go beyond $[-20; 20]^\circ$ from their starting orientation during training, we increase the initial range of test object orientations to $[-20; 20]^\circ$ around the pitch and roll axis}

\paragraph{Naming utterances} We aim to simulate developmentally-relevant statistics of utterances from a caregiver to a \rebut{toddler} during a play session. To achieve this, we source relevant transcripts from the CHILDES database \cite{macwhinney1985child}. Our study focuses specifically on transcripts from native English-speaking children residing in North America and the UK, between the ages of six months and two years. We only keep utterances of caregivers. As we are interested in object-related statements, we curate captions derived from statements identifying objects from the Toys4k dataset \cite{stojanov2021using}. To form general templates, we only keep captions that occur across multiple object categories. After filtering out incomplete utterances and those containing less than three words, as well as a manual review, we retained 820 templates. Each of these templates can be used with any category name without relying on the context of the object. %Examples of captions are shown in \figref{fig:dyadicfull}D.

\subsection{Self-supervised learning of visual representations}\label{sec:ssl}

In this work, we postulate that toddlers learn visual representations that \rebut{1) slowly change over time and 2) align with co-occurring linguistic representations.} To model their learning process, we consider two previously introduced self-supervised loss functions. The first loss aligns embeddings of close-in-time images\rebut{; this entails that} seeing an object from different viewpoints \rebut{elicits} viewpoint invariant object representations \cite{schneider2021contrastive}. The second loss function aligns embeddings of co-occurring images and naming utterances. Since the naming utterance may inform about the category of the object in focus, this provides weak supervision regarding the object category. To implement our learning mechanisms, we use SimCLR \cite{chen2020simple}, a state-of-the-art contrastive learning algorithm. \rebut{We sum up the learning architecture in \figref{fig:dyadicfull}E.}

\paragraph{Contrastive learning through time (CLTT)} At each learning iteration, we sample a mini-batch $\mathcal{X}$ that contains $N$ randomly sampled images $x_i$. For each image $x_i$, we also randomly sample a single successor/predecessor $x_j$ belonging to the same ``play session" as $x_i$. Thus, $x_i$ and $x_j$ always display the same object, but observed through different orientations and positions. In addition, we expect that the active and multi-modal manipulation of the object biases toddlers' attention onto the main object \cite{schroer2023looking,yang2023using}; thus, we apply on images, with probability $0.5$, a center crop of size ranging in $[8; 100]\%$ of the image size. Thereafter, we compute embeddings of images $z^1=g^1(f(x))$ using a feature extractor $f$ and a projection head $g^1$, both implemented as neural networks \cite{chen2020simple}. Finally, for a pair $(z_i,z_j)$, we minimize
\begin{equation}
        l_{\texttt{T}}(z^1_i,z^1_j) = -\log \frac{e^{ {\rm cos}(z^1_i,z^1_j)/\tau}}{\sum_{\substack{z \in \mathcal{Z}^1 \\ z \neq z_i^1}} e^{{\rm cos}(z_i^1, z)/\tau}} \, ,
    \label{eq:simclrtt}
\end{equation}
where $\rm cos$ stands for the cosine similarity, $Z^1$ contains all embeddings $z^1$ and $\tau$ is the temperature hyper-parameter \cite{chen2020simple}. The top part of \eqref{eq:simclrtt} aims to move together image embeddings that belong to the same manipulation session while the bottom part of \eqref{eq:simclrtt} ensures that all image embeddings remain dissimilar.

\paragraph{Multimodal contrastive learning} For each sampled image $x_i$, \rebut{we also sample, if provided by the caregiver, its co-occurring naming utterance $l_i$.} We extract the pre-trained features of the naming utterances with a state-of-art text embedding model $h$ \cite{devlin2019bert}. Then, we compute \rebut{different} embeddings of images $z^2_i=g^2(f(x_i))$ and captions $z^3_i=g^3(h(l_i))$ using projection heads $g^2$, $g^3$ and a text feature extractor $h$ in order to minimize, for each pair $(z^2_i, z^3_i)$:
\begin{equation}
       l_{\texttt{M}}(z^2_i, z^3_i) = -\log \frac{e^{ {\rm cos}(z^2_i,z^3_i)/\tau}}{\sum_{\substack{z \in \mathcal{Z}^2 \\ z \neq z^2_i}} e^{{\rm cos}(z^2_i, z)/\tau}}
    \label{eq:simclrmm}
\end{equation}
where $\mathcal{Z}^2$ \rebut{contains the provided} embeddings $z^2$ and $z^3$. The top part of \eqref{eq:simclrmm} ensures that co-occurring naming utterances and images have similar embeddings while the bottom part prevents all embeddings to collapse into a single vector.

The total loss function for a batch of size $N$ is \rebut{the symmetric and batch-wise sum of \eqref{eq:simclrtt} and \eqref{eq:simclrmm}:}
\begin{equation}
    \frac{1}{2N} \sum_{i=1}^{N} l_{\texttt{T}}(z^1_i,z^1_j) + l_{\texttt{T}}(z^1_j,z^1_i) + l_{\texttt{M}}(z^2_i, z^3_i) + l_{\texttt{M}}(z^3_i, z^2_i). \nonumber
\end{equation}

\subsection{Developmentally-relevant utterance statistics}

%object interactions
To model developmentally-relevant utterance statistics, we extract statistics reported in at-home studies of toddlers' dyadic \rebut{play sessions} \cite{suarez2022joint,schroer2022visual}. First, \cite{schroer2022visual} reports that 56\% of naming events match the visual input of toddlers. Thus, we define the (conditional) probability of naming the manipulated object (versus another object in the background) as $p_{\rm correct}=0.5$. Second, a caregiver approximately provides, on average, one naming utterance related to the object being manipulated per time-extended object manipulation \cite{suarez2022joint} (exactly $\frac{2.54}{2} = 1.27$ in their study). Since our play sessions last for 20 frames, we approximate the probability of naming the manipulated object during a frame as $p_{\rm name}=\frac{1}{20} = 0.05$. \rebut{We assume that consecutive frames are spaced by one second,} making the duration \rebut{of our play sessions correspond to} the average \rebut{of} $20$ seconds reported in \cite{suarez2022joint}. Finally, it allows us to define the probability of naming any object in a frame $p_{\rm sparse} = \frac{p_{\rm name}}{p_{\rm correct}} = 0.1$.
\subsection{Training and evaluation}

\paragraph{Training} We use a ResNet18 \cite{he2016deep} as vision encoder $f$ and, for all projection heads, a fully connected neural network with one hidden layer of size $256$ followed by batch normalization and ReLU activation. For encoding the text, we use a pre-trained BERT \cite{devlin2019bert} with 4 layers, 8 self-attention heads and a hidden size of 512, introduced as BERT-small by \cite{turc2019wellread}.  We train our models for $50$ epochs with the AdamW optimizer, a learning rate of $0.001$ and weight decay of $0.01$. We tested a temperature hyper-parameter $\tau \in \{0.07,0.1,0.5\}$ and found $0.07$ to be the best. \rebut{The rationale behind using a pre-trained text model lies in our research focus on visual representation learning. Preliminary experiments (reported in experiments \secref{sec:mainres}), demonstrate comparable recognition performance, although with a longer convergence time when training a randomly initialized text model concurrently with visual encoding.}

\paragraph{Evaluation} To evaluate if the learned representation supports category recognition, we apply a repeated random sub-sampling cross-validation to split the non-overlapping object instances into $2382$ train ($2/3$ of the total) and $1191$ test objects ($1/3$ of the total). Then, \rebut{we extract images of train and test objects to form the train and test datasets, respectively (see \secref{sec:dataset})}. To evaluate our representation with respect to object instance recognition, we extract novel images of the train objects to build a test dataset. For both object instance and category recognition, we freeze the weights of the vision encoder and train a linear classifier in an online and supervised fashion on top of the latent visual representation \cite{bordes2023towards}. 
\section{Experiments \& Results}\label{sec:results}

We first study whether developmentally-relevant utterance statistics support learning semantic visual representations. Then, we analyze the impact of the sparsity, ambiguity and \rebut{the attention paid to the object's name} on the learning process.

\begin{figure}
    \centering
    \includegraphics[width=\linewidth]{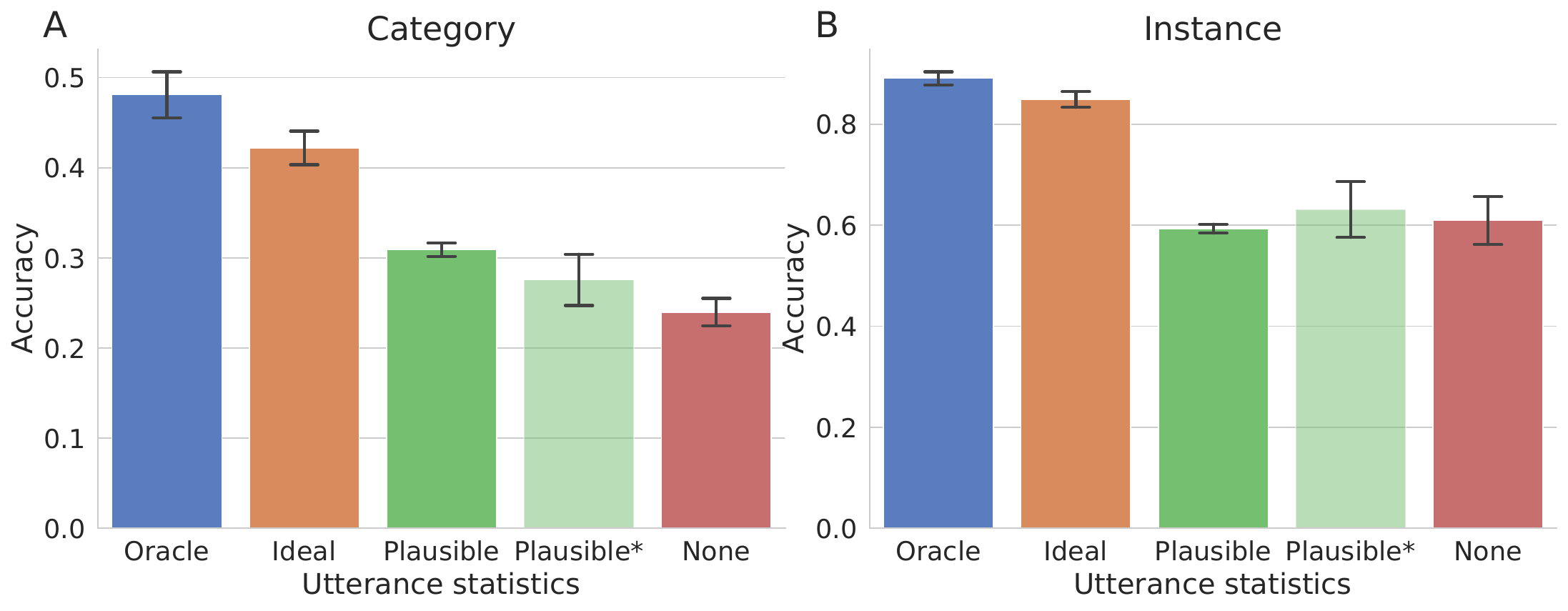}
    \caption{A) Category recognition accuracy and B) object instance recognition accuracy for different settings. \texttt{Oracle} represents supervised learning, while an \texttt{Ideal} caregiver consistently names the correct object. \texttt{Plausible} stands for developmentally-relevant utterance statistics, \texttt{Plausible*} is identical but trains the text encoder from scratch.}
    \label{fig:mainres}
\end{figure}

\subsection{Developmentally relevant utterance statistics improve object representations}\label{sec:mainres}

\begin{figure}
    \centering
    \includegraphics[width=1\linewidth]{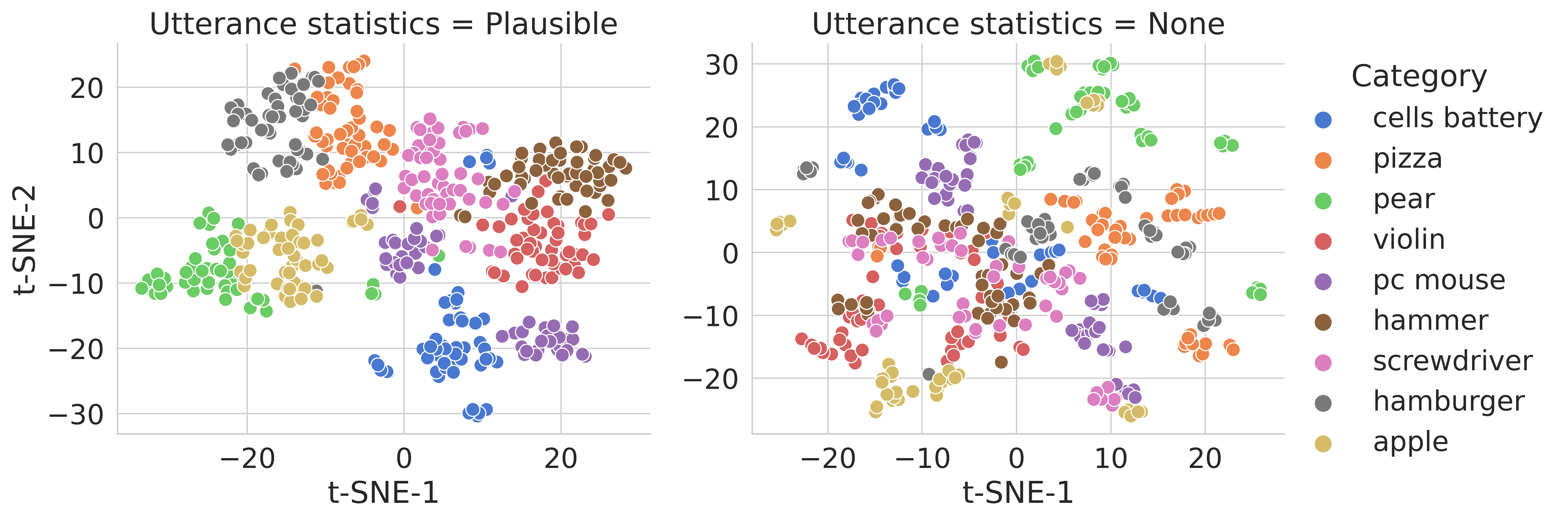}
    \caption{t-SNE visualization of the feature representations extracted by the vision-encoder in different training settings. For better visualization, we show a random subgroup of all classes.}
    \label{fig:tsne}
\end{figure}
%To assess the impact of plausible utterance statistics on the visual representation, \figref{fig:mainres} compares such statistics with two upper-bound baselines (Oracle, Ideal) and a baseline that does not use any utterances (None). 
\rebut{To assess the impact of plausible utterance statistics on the visual representation, \figref{fig:mainres} compares such statistics with two upper-bound baselines: \texttt{Oracle} refers to supervised learning and \texttt{Ideal} refers to a caregiver who always utters sentences containing the correct object. Our baseline (\texttt{None}) does not use any utterances relying solely on the time-contrastive loss for visual representations.}
We observe that our model improves category recognition over not using naming utterances, even though it does not reach the performance of oracles. \rebut{Additionally, we examine how plausible utterance statistics perform when used to train the text representations jointly with the visual representations from scratch over 100 epochs. The model yields similar outcomes, albeit necessitating significantly more training time, as it has not yet reached convergence.}
We employ t-SNE visualization to validate our learned output feature representations. \figref{fig:tsne} shows that the models trained with utterances exhibit better separation than without textual guidance (baseline). 
We conclude that developmentally-relevant imperfect naming utterances can help to build semantic visual representation.

\subsection{Small changes in utterance frequency and ambiguity drastically impact object representations}

\begin{figure}
    \centering
    \includegraphics[width=\linewidth]{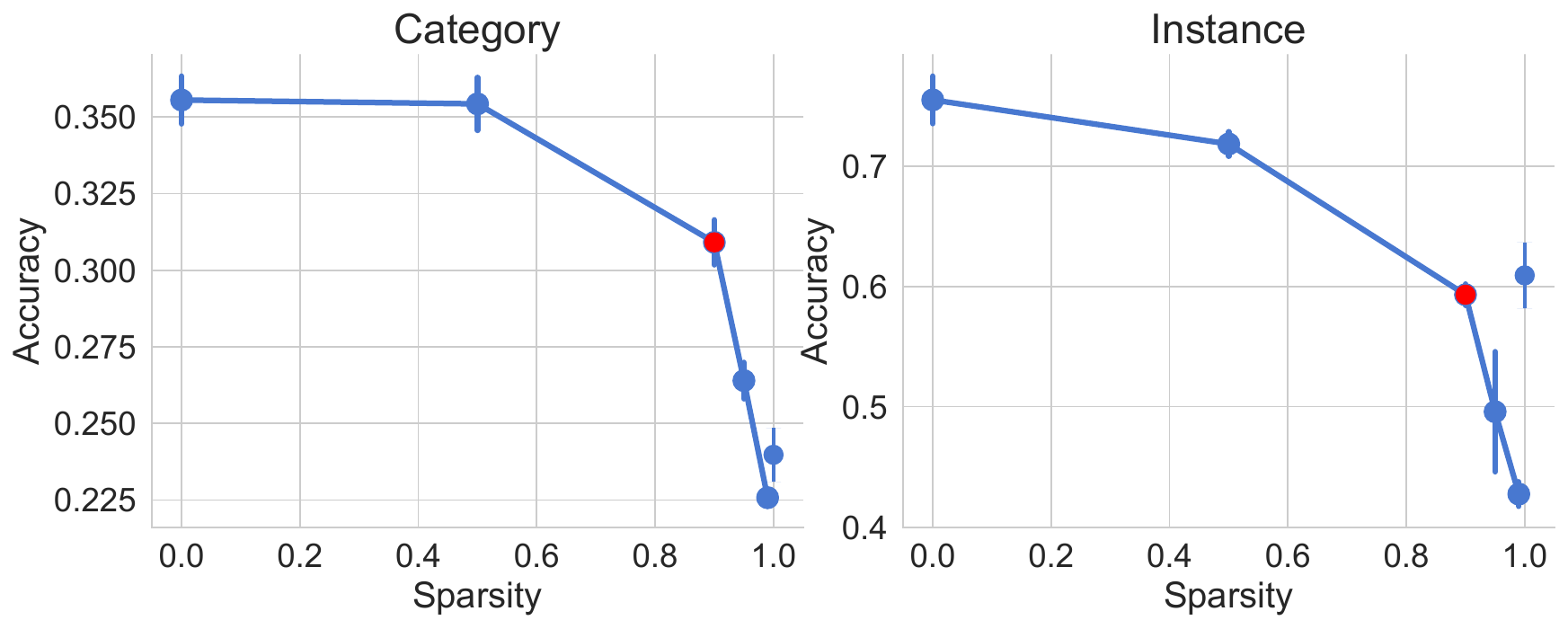}
    \caption{Analysis of the impact of the sparsity parameters on A) category recognition and B) object instance recognition. The red points indicate the developmentally-relevant value.}
    \label{fig:sparse}
\end{figure}

\begin{figure}
    \centering
    \includegraphics[width=\linewidth]{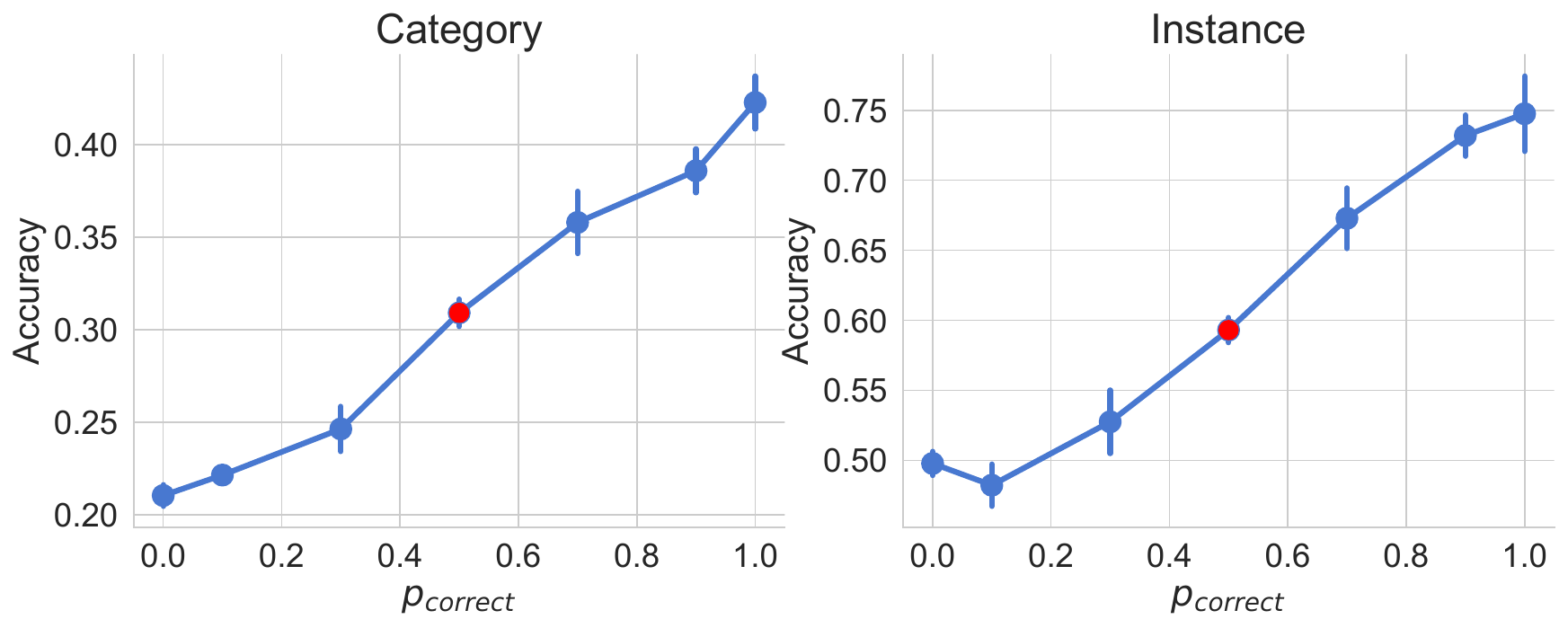}
    \caption{Analysis of the impact of naming ambiguity on A) category recognition and B) instance recognition. A high $p_{\rm correct}$ implies low ambiguity. The red points indicate the developmentally-relevant value.}
    \label{fig:obj}
\end{figure}

To understand how frequency and ambiguity of utterances affect object representations, we systematically vary both factors in \figref{fig:sparse} and \figref{fig:obj}, respectively. We observe that developmentally-relevant values (red points) are close to the steepest point of the function, where small shifts have a high impact on the quality of the learned representation. This suggests that toddler's object learning may be quite sensitive to caregivers' presence and the quality of their utterances. \rebut{We also notice worse instance recognition with very few utterances in comparison to none. We suspect that these rare and ambiguous utterances mostly inject noise into the representation.}

\begin{figure}
    \centering
    \includegraphics[width=\linewidth]{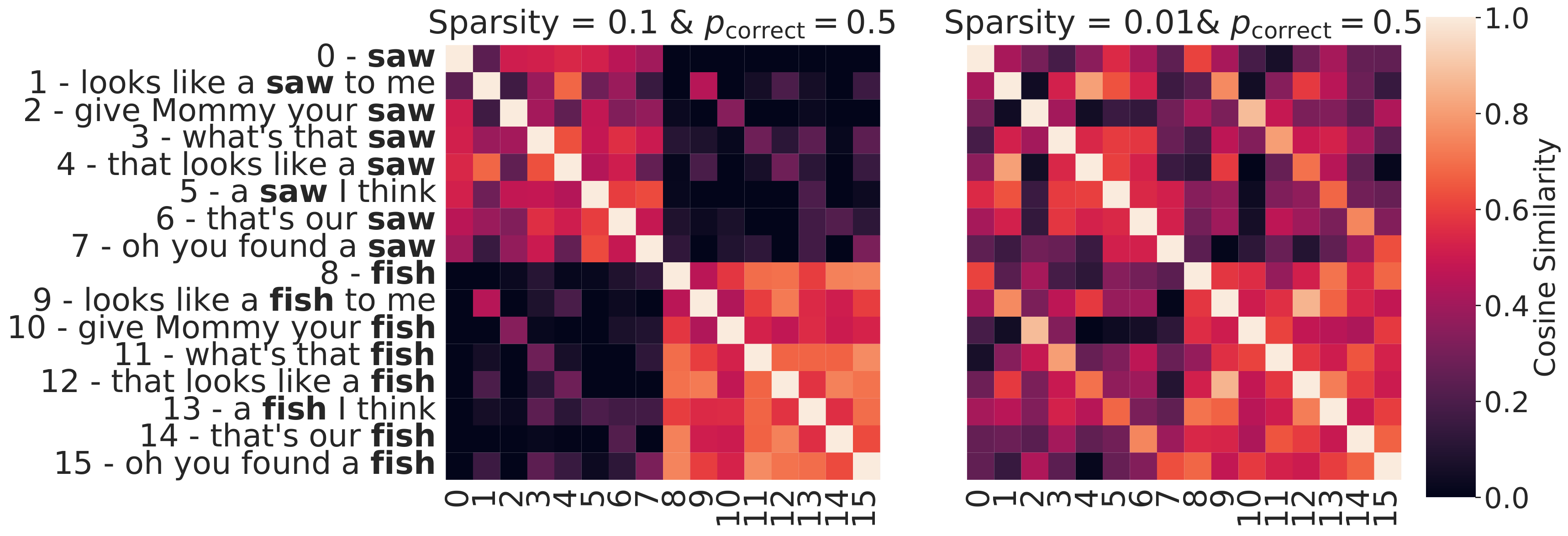}
    \caption{Analysis of the effect of the sparsity parameter and the correct naming probability $p_{\rm correct}$ on the textual representations. The heatmaps show the cosine similarity between the text embeddings of utterances from two random classes. The first element of each class is the raw category name.}
    \label{fig:heatmap}
\end{figure}

\subsection{Attending to object names is crucial for learning good visual representations}
To assess the importance of attention in acquiring semantic visual representations, we present in \figref{fig:heatmap} a comparison of text embeddings across various utterances. Even with the infrequent application of the cross-modal loss, the model demonstrates the capability to group utterances sharing the same category name, clearly separating them from other utterance classes (as illustrated by the blocks in \figref{fig:heatmap}). Increasing the degree of textual guidance by small increments yields improved class separation. In addition, more good naming utterances amplify linguistic attention on the name of the object being held, thereby allowing to better guide the visual representation. 
When considered alongside the accuracy values of classification experiments, this observation strongly suggests that increased textual attention to the object name is crucial for learning object representations.

\section{Conclusion}

\rebut{We proposed a computational model of the development of toddlers object representations during dyadic play and investigated if and how caregivers' utterances affect learned visual representations.} We found that realistic utterance statistics elicit more semantic visual representations and the quality of these representations is sensitive to small shifts in the statistics of utterances.
Our analysis also revealed that realistic statistics of utterances \rebut{lead} the model to focus on the object name within an utterance, suggesting that visual and language representations reciprocally affect each other.

\rebut{The main limitation of our work} is the gap between the actual experiences of toddlers and our simplified dataset. For instance, our ``play sessions" are short and disconnected, as there is no notion of temporal proximity between different sessions and our captions represent a small subset of the space of possible utterances. In addition, hands, caregivers' visual cues and the touch modality have been completely ignored. In the future, we plan to validate our findings on a dataset of real-world videos \rebut{extracted} from head-mounted cameras during dyadic play.

\section*{ACKNOWLEDGMENT}
We gratefully acknowledge support from GENCI–IDRIS (Grant 2022-AD011014008) for providing computing and data-processing resources needed for this work. Additional support was received by the Deutsche Forschungsgemeinschaft (DFG project 5368 ``Abstract REpresentations in Neural Architectures (ARENA)''), as well as the project ``The Adaptive Mind'' funded by the Excellence Program of the Hessian Ministry of Higher Education, Science, Research and Art (HMWK). Jochen Triesch was supported by the Johanna Quandt foundation.

%%%%%%%%% REFERENCES
{\small
\bibliographystyle{ieee_fullname}
\bibliography{main}

\begin{thebibliography}{10}\itemsep=-1pt

\bibitem{aubrettime}
Arthur Aubret, Markus~R. Ernst, C{\'e}line Teuli{\`e}re, and Jochen Triesch.
\newblock Time to augment self-supervised visual representation learning.
\newblock In {\em The Eleventh International Conference on Learning
  Representations}, 2023.

\bibitem{aubret2022toddler}
Arthur Aubret, C{\'e}line Teuli{\`e}re, and Jochen Triesch.
\newblock Toddler-inspired learning induces hierarchical object
  representations.
\newblock In {\em IEEE ICDL-Sensorimotor Interaction, language, and Embodiement
  of Symbols (SMILES) workshop}, 2022.

\bibitem{bambach2016active}
Sven Bambach, David~J Crandall, Linda~B Smith, and Chen Yu.
\newblock Active viewing in toddlers facilitates visual object learning: An
  egocentric vision approach.
\newblock In {\em CogSci}, 2016.

\bibitem{bordes2023towards}
Florian Bordes, Randall Balestriero, and Pascal Vincent.
\newblock Towards democratizing joint-embedding self-supervised learning.
\newblock {\em arXiv preprint arXiv:2303.01986}, 2023.

\bibitem{chen2020simple}
Ting Chen, Simon Kornblith, Mohammad Norouzi, and Geoffrey Hinton.
\newblock A simple framework for contrastive learning of visual
  representations.
\newblock In {\em International conference on machine learning}, pages
  1597--1607. PMLR, 2020.

\bibitem{clerkin2022real}
Elizabeth~M Clerkin and Linda~B Smith.
\newblock Real-world statistics at two timescales and a mechanism for infant
  learning of object names.
\newblock {\em Proceedings of the National Academy of Sciences},
  119(18):e2123239119, 2022.

\bibitem{desai2021virtex}
Karan Desai and Justin Johnson.
\newblock Virtex: Learning visual representations from textual annotations.
\newblock In {\em Proceedings of the IEEE/CVF conference on computer vision and
  pattern recognition}, pages 11162--11173, 2021.

\bibitem{devlin2019bert}
Jacob Devlin, Ming-Wei Chang, Kenton Lee, and Kristina Toutanova.
\newblock {BERT}: Pre-training of deep bidirectional transformers for language
  understanding.
\newblock In {\em Proceedings of the 2019 ACL}, pages 4171--4186, Minneapolis,
  Minnesota, June 2019. Association for Computational Linguistics.

\bibitem{diesendruck2003specific}
Gil Diesendruck and Paul Bloom.
\newblock How specific is the shape bias?
\newblock {\em Child development}, 74(1):168--178, 2003.

\bibitem{feichtenhofer2021large}
Christoph Feichtenhofer, Haoqi Fan, Bo Xiong, Ross Girshick, and Kaiming He.
\newblock A large-scale study on unsupervised spatiotemporal representation
  learning.
\newblock In {\em Proceedings of the IEEE/CVF Conference on Computer Vision and
  Pattern Recognition}, pages 3299--3309, 2021.

\bibitem{franzius2011invariant}
Mathias Franzius, Niko Wilbert, and Laurenz Wiskott.
\newblock Invariant object recognition and pose estimation with slow feature
  analysis.
\newblock {\em Neural computation}, 23(9):2289--2323, 2011.

\bibitem{gan2020threedworld}
Chuang Gan, Jeremy Schwartz, Seth Alter, Damian Mrowca, Martin Schrimpf, James
  Traer, Julian De~Freitas, Jonas Kubilius, Abhishek Bhandwaldar, Nick Haber,
  et~al.
\newblock Threedworld: A platform for interactive multi-modal physical
  simulation.
\newblock {\em arXiv preprint arXiv:2007.04954}, 2020.

\bibitem{gershkoff2004shape}
Lisa Gershkoff-Stowe and Linda~B Smith.
\newblock Shape and the first hundred nouns.
\newblock {\em Child development}, 75(4):1098--1114, 2004.

\bibitem{gopnik1987development}
Alison Gopnik and Andrew Meltzoff.
\newblock The development of categorization in the second year and its relation
  to other cognitive and linguistic developments.
\newblock {\em Child development}, pages 1523--1531, 1987.

\bibitem{harwath2018jointly}
David Harwath, Adria Recasens, D{\'\i}dac Sur{\'\i}s, Galen Chuang, Antonio
  Torralba, and James Glass.
\newblock Jointly discovering visual objects and spoken words from raw sensory
  input.
\newblock In {\em Proceedings of the European conference on computer vision
  (ECCV)}, pages 649--665, 2018.

\bibitem{he2016deep}
Kaiming He, Xiangyu Zhang, Shaoqing Ren, and Jian Sun.
\newblock Deep residual learning for image recognition.
\newblock In {\em Proceedings of the IEEE conference on computer vision and
  pattern recognition}, pages 770--778, 2016.

\bibitem{jones2003late}
Susan~S Jones.
\newblock Late talkers show no shape bias in a novel name extension task.
\newblock {\em Developmental Science}, 6(5):477--483, 2003.

\bibitem{kraebel2006three}
Kimberly~S Kraebel and Peter~C Gerhardstein.
\newblock Three-month-old infants’ object recognition across changes in
  viewpoint using an operant learning procedure.
\newblock {\em Infant Behavior and Development}, 29(1):11--23, 2006.

\bibitem{li2010unsupervised}
Nuo Li and James~J DiCarlo.
\newblock Unsupervised natural visual experience rapidly reshapes
  size-invariant object representation in inferior temporal cortex.
\newblock {\em Neuron}, 67(6):1062--1075, 2010.

\bibitem{macwhinney1985child}
Brian MacWhinney and Catherine Snow.
\newblock The child language data exchange system.
\newblock {\em Journal of child language}, 12(2):271--295, 1985.

\bibitem{nelson2000young}
Deborah G~Kemler Nelson, Anne Frankenfield, Catherine Morris, and Elizabeth
  Blair.
\newblock Young children's use of functional information to categorize
  artifacts: Three factors that matter.
\newblock {\em Cognition}, 77(2):133--168, 2000.

\bibitem{orhan2020self}
Emin Orhan, Vaibhav Gupta, and Brenden~M Lake.
\newblock Self-supervised learning through the eyes of a child.
\newblock In H. Larochelle, M. Ranzato, R. Hadsell, M.F. Balcan, and H. Lin,
  editors, {\em Advances in Neural Information Processing Systems}, volume~33,
  pages 9960--9971. Curran Associates, Inc., 2020.

\bibitem{pereira2010early}
Alfredo~F Pereira, Karin~H James, Susan~S Jones, and Linda~B Smith.
\newblock Early biases and developmental changes in self-generated object
  views.
\newblock {\em Journal of vision}, 10(11):22--22, 2010.

\bibitem{radford2021learning}
Alec Radford, Jong~Wook Kim, Chris Hallacy, Aditya Ramesh, Gabriel Goh,
  Sandhini Agarwal, Girish Sastry, Amanda Askell, Pamela Mishkin, Jack Clark,
  et~al.
\newblock Learning transferable visual models from natural language
  supervision.
\newblock In {\em International conference on machine learning}, pages
  8748--8763. PMLR, 2021.

\bibitem{schneider2021contrastive}
Felix Schneider, Xia Xu, Markus~R Ernst, Zhengyang Yu, and Jochen Triesch.
\newblock Contrastive learning through time.
\newblock In {\em SVRHM 2021 Workshop@ NeurIPS}, 2021.

\bibitem{schroer2022visual}
Sara~E Schroer, Ryan~E Peters, Alyssa Yarbrough, and Chen Yu.
\newblock Visual attention and language exposure during everyday activities: an
  at-home study of early word learning using wearable eye trackers.
\newblock In {\em Proceedings of the Annual Meeting of the Cognitive Science
  Society}, volume~44, 2022.

\bibitem{schroer2023looking}
Sara~E Schroer and Chen Yu.
\newblock Looking is not enough: Multimodal attention supports the real-time
  learning of new words.
\newblock {\em Developmental Science}, 26(2):e13290, 2023.

\bibitem{smith2002object}
Linda~B Smith, Susan~S Jones, Barbara Landau, Lisa Gershkoff-Stowe, and Larissa
  Samuelson.
\newblock Object name learning provides on-the-job training for attention.
\newblock {\em Psychological science}, 13(1):13--19, 2002.

\bibitem{soska2008development}
Kasey~C Soska and Scott~P Johnson.
\newblock Development of three-dimensional object completion in infancy.
\newblock {\em Child development}, 79(5):1230--1236, 2008.

\bibitem{stojanov2021using}
Stefan Stojanov, Anh Thai, and James~M. Rehg.
\newblock Using shape to categorize: Low-shot learning with an explicit shape
  bias.
\newblock In {\em Proceedings of the IEEE/CVF Conference on Computer Vision and
  Pattern Recognition (CVPR)}, pages 1798--1808, June 2021.

\bibitem{suarez2022joint}
Catalina Suarez-Rivera, Jacob~L Schatz, Orit Herzberg, and Catherine~S
  Tamis-LeMonda.
\newblock Joint engagement in the home environment is frequent, multimodal,
  timely, and structured.
\newblock {\em Infancy}, 27(2):232--254, 2022.

\bibitem{sullivan2021saycam}
Jessica Sullivan, Michelle Mei, Andrew Perfors, Erica Wojcik, and Michael~C
  Frank.
\newblock Saycam: A large, longitudinal audiovisual dataset recorded from the
  infant’s perspective.
\newblock {\em Open mind}, 5:20--29, 2021.

\bibitem{tsutsui2020computational}
Satoshi Tsutsui, Arjun Chandrasekaran, Md~Alimoor Reza, David Crandall, and
  Chen Yu.
\newblock A computational model of early word learning from the infant's point
  of view.
\newblock {\em arXiv preprint arXiv:2006.02802}, 2020.

\bibitem{turc2019wellread}
Iulia Turc, Ming-Wei Chang, Kenton Lee, and Kristina Toutanova.
\newblock Well-read students learn better: On the importance of pre-training
  compact models, 2019.

\bibitem{vong2022cross}
Wai~Keen Vong and Brenden~M Lake.
\newblock Cross-situational word learning with multimodal neural networks.
\newblock {\em Cognitive science}, 46(4):e13122, 2022.

\bibitem{vong2021cross}
Wai~Keen Vong, Emin Orhan, and Brenden Lake.
\newblock Cross-situational word learning from naturalistic headcam data.
\newblock In {\em 34th CUNY Conference on Human Sentence Processing}, 2021.

\bibitem{wiskott2002slow}
Laurenz Wiskott and Terrence~J Sejnowski.
\newblock Slow feature analysis: Unsupervised learning of invariances.
\newblock {\em Neural computation}, 14(4):715--770, 2002.

\bibitem{wood2018development}
Justin~N Wood and Samantha~MW Wood.
\newblock The development of invariant object recognition requires visual
  experience with temporally smooth objects.
\newblock {\em Cognitive Science}, 42(4):1391--1406, 2018.

\bibitem{yang2023using}
Jane Yang, Linda Smith, David Crandall, and Chen Yu.
\newblock Using manual actions to create visual saliency: an outside-in
  solution to sustained attention and joint attention.
\newblock In {\em Proceedings of the Annual Meeting of the Cognitive Science
  Society}, 2023.

\bibitem{yu2021infant}
Chen Yu, Yayun Zhang, Lauren~K Slone, and Linda~B Smith.
\newblock The infant’s view redefines the problem of referential uncertainty
  in early word learning.
\newblock {\em Proceedings of the National Academy of Sciences},
  118(52):e2107019118, 2021.

\end{thebibliography}
}

\end{document}